\documentclass{article} 
\usepackage[final]{neurips_data_2024}

\usepackage{soul}
\usepackage{wrapfig}

\usepackage{listings}

\usepackage{microtype}
\usepackage{hyperref}
\usepackage{url}
\usepackage{booktabs}
\usepackage{authblk}
\usepackage{comment}
\usepackage{todonotes}
\usepackage{amsthm}
\usepackage{mdframed}
\usepackage{bbm}

\usepackage[most]{tcolorbox}

\usepackage{bbding}
\usepackage{cleveref}

\theoremstyle{definition}

\newtheorem*{prob*}{The Key Problem}
\surroundwithmdframed{prob*}

\DeclareMathOperator{\argmax}{argmax}

\usepackage{listings}

\usepackage{relsize}
\newcommand{\lmevaluationharness}{\texttt{Language Model Evaluation Harness}}
\newcommand{\evalharness}{\texttt{lm-eval}}

\newcommand\hailey[1]{\textcolor{red}{Hailey: #1}}

\makeatletter
\renewcommand\AB@affilsepx{, \protect\Affilfont}

\makeatletter
\renewcommand\AB@affilsepx{, \protect\Affilfont}

\newcommand{\affilbreak}[1]{%
    \renewcommand\AB@affilsepx{\\[\baselineskip]\protect\Affilfont}
    #1
    \renewcommand\AB@affilsepx{, \protect\Affilfont}
}
\makeatother

\title{Lessons from the Trenches on\\Reproducible Evaluation of Language Models}

\author[1*]{Stella Biderman}
\author[1*]{Hailey Schoelkopf}
\author[1*]{Lintang Sutawika}
\author[1$\dag$]{\authorcr Baber Abbasi}
\author[7$\dag$]{Jessica Zosa Forde}
\author[1$\dag$]{Leo Gao}
\author[2$\dag$]{Jonathan Tow}
\author[3]{\authorcr Alham Fikri Aji}
\author[4]{Pawan Sasanka Ammanamanchi}
\author[1]{Sidney Black}
\author[5]{Jordan Clive}
\author[1]{Anthony DiPofi}
\author[6]{Julen Etxaniz}
\author[1]{Benjamin Fattori}
\author[8]{Charles Foster}
\author[9]{Jeffrey Hsu}
\author[10]{Mimansa Jaiswal}
\author[11]{Wilson Y. Lee}
\author[3,12]{Haonan Li}
\author[13]{Charles Lovering}
\author[14]{Niklas Muennighoff}
\author[7]{Ellie Pavlick}
\author[1,15]{Jason Phang}
\author[1]{Aviya Skowron}
\author[16]{Samson Tan}
\author[17]{Xiangru Tang}
\author[7]{Kevin A. Wang}
\author[18]{Genta Indra Winata}
\author[19]{Fran\c{c}ois Yvon}
\author[20]{Andy Zou}

\affil[1]{EleutherAI}
\affil[2]{Stability AI}
\affil[3]{MBZUAI}
\affil[4]{IIIT Hyderabad}
\affil[5]{Chattermill AI}
\affil[6]{HiTZ Center - Ixa, UPV/EHU}
\affil[7]{Brown University}
\affil[8]{Finetune}
\affil[9]{Ivy Natal}
\affil[10]{University of Michigan}
\affil[11]{HubSpot}
\affil[12]{LibrAI}
\affil[13]{Kensho}
\affil[14]{Contextual AI}
\affil[15]{New York University}
\affil[16]{Amazon}
\affil[17]{Yale University}
\affil[18]{HKUST}
\affil[19]{Sorbonne University}
\affilbreak{\affil[20]{CMU}}

\affil[*]{Lead authors}
\affil[$\dag$]{Second authors}

\begin{document}

\maketitle

\begin{abstract}

Reliable evaluation of language models (LMs) remains an open challenge. Researchers and engineers face methodological issues such as the sensitivity of models to evaluation setup, difficulty of proper comparisons across methods, and the lack of reproducibility and transparency. Evaluation difficulties are exacerbated by the fracturing and siloing of information about conventions and common practices. In this paper we draw on three years of experience in evaluating large language models (LMs) as developers of the popular~\lmevaluationharness~(\evalharness) \citep{eval-harness} framework to provide guidance and lessons for the field moving forward. We document a variety of challenges faced by practitioners and provide concrete instances where these challenges or the absence of best practices have come into effect. We make recommendations to the field for improving evaluation rigor and confidence, and attempt to codify much of the tacit or folk knowledge surrounding LM evaluation, for a solid ground to move forward.
\end{abstract}

\section{Introduction}

Efforts to progress the study of LMs and machine learning hinge upon the communication of evaluation results presented to support new ideas and conclusions. Improper evaluation practices can lead to skewed performance comparisons \citep{card2020little} which may influence the direction of future research and the adoption of new methods by the community \citep{dehghani2021benchmark}, or lead to adverse effects from deploying suboptimal or harmful models \citep{bender-friedman-2018-data} on tasks for which they are ill-suited \citep{Raji_2022}. However, evaluation in LMs is fraught with difficulties, both due to problems inherent in benchmarking across fields and due to specific complications or challenges introduced by LMs. As the study of LMs and their typical uses have rapidly shifted over the past years, evaluation practices have been left by the wayside, resulting in an evaluation landscape that is quickly changing, with accepted common practices either obscurely or not at all formally documented, and a fractured ecosystem.

In this paper we attempt to rectify this situation by documenting the current state of evaluations and our lessons learned that have been especially beneficial in obtaining useful and rigorous findings. Drawing on our experience building and maintaining \evalharness~\citep{eval-harness}, a flexible LM evaluation library, we identify common challenges, recommend best practices, and share how we have operationalized these best practices in our work. Unlike other evaluation libraries which prescribe specific benchmarks \citep{liang2023holistic, opencompass}, \evalharness~is intentionally non-opinionated about \textit{what} to evaluate. It is an \textit{orchestration} library: its role is to make it possible to run any benchmark on any model reliably and reproducibly, not to dictate which benchmarks matter. We are, however, opinionated about \textit{how} a given evaluation should be run---each task ships with a carefully chosen default configuration reflecting community best practices---while deliberately making it easy for power users to customize any aspect of their evaluation setup. The goal is to guide users toward sound methodology by default while enabling arbitrary flexibility for advanced users.

In this way, \evalharness~seeks to serve as \textbf{research infrastructure}---a resource for researching evaluation, locating de facto best practices, and enhancing reproducibility. The design of \evalharness~has evolved over time to reflect the needs of the open-source community and our evolving understanding of best practices; we discuss specific ways we have modified the library to address the challenges we describe throughout this paper.


Our goals are threefold:

\begin{enumerate}
    \item We collect, synthesize, and systematically present challenges commonly faced when conducting LM evaluations, coupled with concrete instances where these challenges have surfaced. We seek to formalize much of the tacit ``folk knowledge'' held within the field surrounding evaluation practices and implementation details.
    \item We provide recommendations for best practices in LM evaluations, informed by these practical challenges.
    \item We document how we have incorporated these best practices into newer versions of our library ~\evalharness. We intend this to provide an example of operationalizing our stated best practices and ease adoption of our recommendations.
\end{enumerate}

\section{Why is Evaluating Language Models Hard?}
\label{sec:background}

Before turning to specific lessons, we outline three under-discussed aspects of language model evaluation that make it uniquely difficult to do well: a fundamental methodological challenge we term the Key Problem, social dynamics and incentive structures that work against rigorous evaluation, and a rapidly shifting landscape that can render today's best practices obsolete within a year or two. These challenges underpin much of the discussion in this paper.

\subsection{The Key Problem}\label{sec:key-problem}

A useful framing which shapes all of LM evaluation is a concept which we term the \textbf{Key Problem}:

\begin{prob*}
	When evaluating language models, there can be many semantically equivalent but syntactically different ways of expressing the same idea. In an ideal world, to assess correctness we would have a way to automatically detect when two sentences express the same content but in different words. However, our best automated tools for determining whether two sentences are semantically equivalent are the very models we are seeking to evaluate.
\end{prob*}

Many problems with, and common approaches to, LM evaluation are designed around this challenge. Responses to the Key Problem generally fall into three broad categories: \textbf{converting the open-domain problem into a closed-domain one} where correctness can be determined mechanically, using \textbf{model-based grading}, or relying on \textbf{human labor}. Each approach has significant trade-offs.

\paragraph{Closed-domain conversion.} The Key Problem can be sidestepped by artificially restricting the answer space so that correctness can be checked mechanically. The most prevalent way to achieve this is to reframe questions as multiple choice problems (Appendix \ref{app:formal-measurements}), with a single gold target answer and a finite, static set of possible responses \citep{hendrycks2020measuring, Bigbench, Lievin2022CanLL, lin-etal-2022-truthfulqa, robinson2023leveraging, holtzman2022surface}. Another approach is structured generation, in which the model's output is constrained at decoding time to match a specified format, reducing variance from prompt sensitivity and enabling more reliable answer extraction \citep{fourrier2024structured}. Alternatively, when a reference answer is known, one can perform string-matching approaches heuristically to determine whether the model's answer matches the ground truth \citep{dua2019drop,joshi2017triviaqa,hendrycks2021measuring}. In rare cases where we have access to a \textit{practical verifier} \citep{lewkowycz2022solving, chen2021codex} it can be used to check correctness directly.


\paragraph{Model-based grading.} A second approach uses automated metrics to directly approximate a solution to the Key Problem by measuring the distance from a generated response to a gold-standard one. Heuristic metrics such as BLEU \citep{papineni-etal-2002-bleu} and ROUGE \citep{lin2004rouge} count the n-gram overlap between the two texts. These metrics offer notable advantages in that they are (theoretically) fully reproducible, far easier and cheaper to compute, and can avoid some of the issues faced by human studies \citep{wei2021statistical, freitag-etal-2021-experts, amidei-etal-2020-identifying}. Nevertheless these heuristic-based metrics have flaws \citep{callison-burch2006re-evaluating} and underdocumented hyperparameters can present reproducibility challenges \citep{marie-etal-2021-scientific}. More recently, model-based metrics have gained momentum through evaluation methods that leverage large language models as a grader \citep{kim2024prometheus, wang2024pandalm, liu-etal-2023-g} --- especially as proxies for human preference evaluation \citep{DBLP:journals/corr/abs-2306-05685} --- but these are known to be flawed \citep{Wang2023LargeLM, shen-etal-2023-large, zeng2024evaluating, Hu2024AreLE, liu2023benchmarking, Chen2024HumansOL} and suffer from similar issues.

\paragraph{Human labor.} Finally, the Key Problem could in principle be addressed by having expert human annotators score model responses for correctness. However, performing accurate human studies is time-consuming and expensive. Additionally, human assessments can be flawed and biased, especially for complex judgments such as factual correctness \citep{hosking2024human, xu-etal-2023-critical, wu2023style}. Expertly trained human judgment can alleviate these issues but is inherently non-scalable.

The Key Problem is an inherent part of language model evaluation. There is no ``correct'' answer, only a best answer in the context that an evaluation is being done. In this paper we focus on MCQA evaluations because they are the approach we have the most experience with and are the most common approach in the literature. We believe that much of our analysis will apply to other approaches, but urge experts in those areas to document and promote their thoughts on best practices in their areas as well.


\subsection{Social Dynamics of Evaluation}\label{sec:social-dynamics}

Beyond its technical difficulties, language model evaluation operates within an incentive landscape that actively works against rigor. Model developers face strong pressure to present their models as the best. While this has always been an issue with peer review in AI, in recent years it has been massively exacerbated by the need to justify the extreme levels of capital investment required to build frontier models, to attract customers, and to maintain competitive positioning. Evaluations are not just scorecards: they are advertisements.

The most fundamental barrier to independent evaluation is access. Some of the allegedly most capable models in the world have never been publicly released in any form. Google's PaLM \citep{chowdhery2023pathways,anil2023palm} and models built on it \citep{lewkowycz2022solving,singhal2023medpalm2}, Baidu's ERNIE~4.0, ByteDance's Doubao, and Tencent's Hunyuan all claim to be very powerful models but there is no way to get access to them. For models like these, the only evaluation numbers that exist are the ones their developer choses to publish.
 
Models that are available through APIs present a subtler problem. While API access does enable independent evaluation in principle, many APIs do not expose output log-probabilities which are required for many standard evaluation benchmarks (See \Cref{app:formal-measurements} for details). This forces evaluators to rely on generation-based approaches, which introduce additional variance from prompt formatting and answer extraction, or to simply trust the developer's self-reported numbers on tasks that require log-probabilities. Running comprehensive benchmark suites through APIs is also expensive, and these costs fall disproportionately on independent researchers---precisely the parties whose assessments would be most credible by virtue of their independence. The field has, in effect, adopted a norm where model developers grade their own homework.

Even when independent evaluation is possible, there is pressure to cherry-pick favorable tasks, choose flattering prompting strategies, or report single-run point estimates that obscure uncertainty. The lack of standardized reporting requirements (discussed further in Section~\ref{sec:Evaluation Details are Lost in Communication}) makes it easy to present numbers that are technically accurate but misleading. As we discuss in Section~\ref{sec:Evaluations are Not Just Numbers}, these dynamics are severe enough that the Benchmark Lottery has become a real phenomenon, with benchmark selection itself becoming a target for optimization rather than a means of measuring genuine progress.

\subsection{Fast-changing Progress and Conventions}\label{sec:fast-changing}

\begin{figure}[ht]
    \centering
    \includegraphics[width=\linewidth]{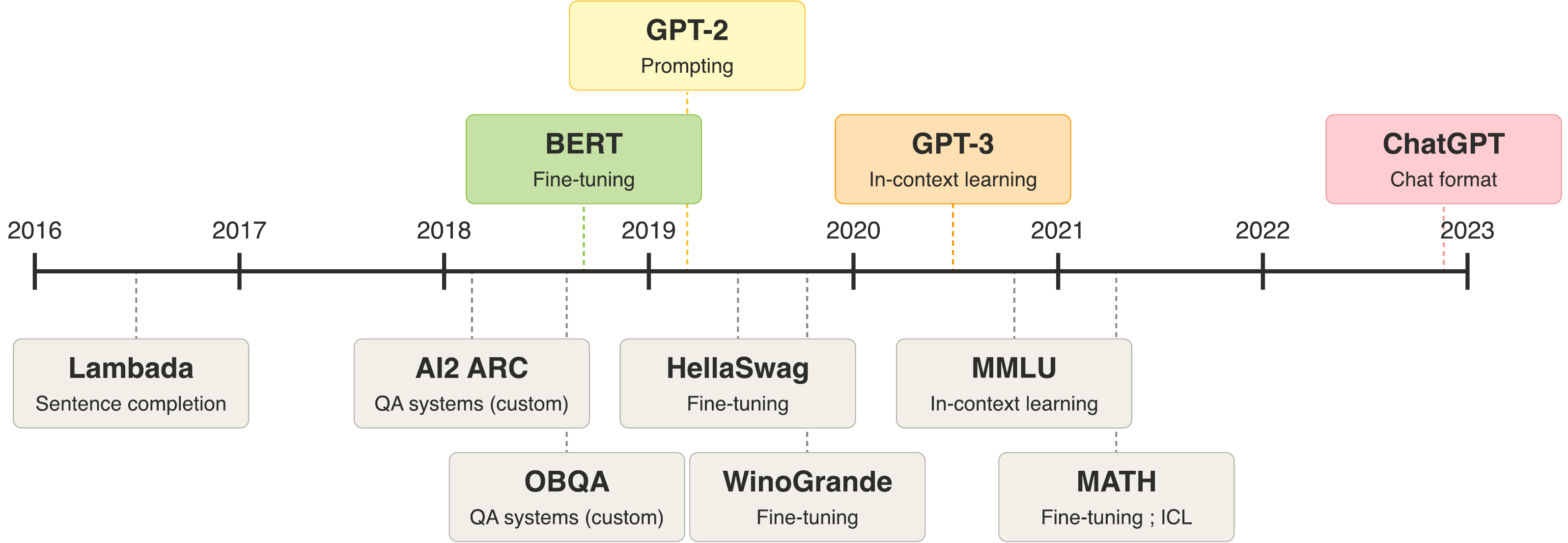}
    \caption{A timeline showing the relative release dates of a selection of notable benchmarks used to evaluate LMs, as compared to the release dates of BERT \citep{devlin2018bert}, GPT-2 \citep{Radford2019Language}, GPT-3 \citep{brown2020language}, and ChatGPT, used as approximate stand-ins for shifts in how the community uses and therefore evaluates LMs. Common practice for evaluating autoregressive language models today diverges from the method described in the paper for all listed tasks except MMLU and MATH.}
    \label{fig:benchmark-timelines}
\end{figure}

Due to the time-consuming nature of developing good benchmarks and the rapid pace of change in NLP research in the past decade, many widely used language model evaluation benchmarks do not represent the current paradigm of how language models are trained: This has two major impacts:
\begin{enumerate}
    \item Benchmarks are being used for purposes outside of their original design, with unclear validity in these new scenarios: for example, a large number of benchmarks \citep{wang2019superglue,wang2019glue} have been built around fine-tuning on a known training set and closed space of labels, but are used today for open-domain language models.
    \item There is no ``ground-truth'' implementation from the original benchmark authors, and many popular benchmarks \citep{paperno2016lambada, clark2018think, zellers2019hellaswag} that have been ``retrofitted'' for compatibility with autoregressive LMs. In the absence of a clear standard, the community's methodology for evaluating on these benchmarks may be fragmented or undocumented.
\end{enumerate}

To illustrate the effects of this development timeline, \autoref{fig:benchmark-timelines} shows how many prominent LM benchmarks were designed prior to shifts such as in-context learning and chat interaction, and therefore were not designed to take these approaches into account. This can affect validity in unforeseen ways.

Aside from validity concerns, it creates a challenge in which popularly used benchmarks may have no clear ``ground-truth'' evaluation code or settings for use in these new areas, requiring practitioners to derive their own design choices and fracturing practices among the community.

\section{Evaluation Details are Lost in Communication}\label{sec:Evaluation Details are Lost in Communication}

The typical benchmark life cycle is a game of telephone--a benchmark is first designed by researchers and released to the community alongside, typically, code for evaluating the benchmark. Next, for the benchmark to see wide usage and community adoption, researchers beyond this initial group must use and run the benchmark in their own work. This typically requires a \textbf{reimplementation step}, where a variety of new models or codebases must interface with the new benchmark in order for it to be widely run. This process requires a process of repeatedly communicating and adapting the benchmark's implementation by new researchers.

In the course of such transmission, it is all too easy for key facts and methodological details to be obscured or fail to be properly adapted. Knowledge of how to perform a benchmark can be implicit or fractured across many locations, resulting in comparisons that may be irreproducible or not 1-to-1, and reducing trust in evaluations run by other researchers. With this in mind, we will explore how this creates challenges for LM benchmarking, and how to move toward resolving it.



\subsection{Challenges}\label{sec:communication-challenges}

\paragraph{``Minor'' Implementation Details Matter}

While one might be tempted to pass off these minor implementation or methodological details as simply trivialities, they in fact significantly affect numerical scores and the conclusions drawn. LMs are not robust, often in seemingly unintuitive ways: for example, the precise string used for a given prompt, even down to whitespace usage, can significantly impact performance \citep{weber2023icl, formatspreadbenchmarks, mizrahi2024state, alzahrani2024benchmarks, webson-pavlick-2022-prompt}. Even the choice of few-shot examples can also significantly change performance \citep{lu-etal-2022-fantastically, min2022rethinking}. In Table \ref{tab:prompt-sensitivity} we show scores for several models on ARC (Challenge) \citep{clark2018think} and MMLU \citep{hendrycks2020measuring} for two different common prompting styles, and see that \textbf{performance varies dramatically with the prompt} (in some cases, $>20\%$), and \textbf{which prompt style works best varies by model}. Thus, the selection of prompt affects the \textit{model rankings and research conclusions} produced in addition to numeric scores themselves \citep{alzahrani2024benchmarks}. These results appear to be consistent with similar analysis done in \citet{gu2024olmes}. Further details can be found in Appendix \ref{app:case-studies-prompt}.

\begin{table}[!h]
\label{tab:prompt-sensitivity}
\caption{Comparison of 0-shot model performance (\texttt{acc}) for several pretrained LMs \citep{gpt-neox-20b, touvron2023llama2, penedo2023refinedweb, jiang2023mistral, jiang2024mixtral} on ARC (Challenge subset) and MMLU across two commonly used prompt styles. }
\centering
\begin{tabular}{ccccccc}
\toprule
            & 
            \multicolumn{2}{c}{ARC (Challenge)} & \multicolumn{2}{c}{MMLU}  \\
            & Cloze      & MMLU-style   & Hybrid & MMLU-style     \\
\midrule
GPT-NeoX-20B  
& \textbf{38.0} $\pm$ \textbf{2.78} \%    & $26.6 \pm 2.53$\%      & \textbf{27.6} $\pm$ \textbf{0.74}\%      & $24.5 \pm 0.71$\%  \\
Llama-2-7B 
& \textbf{43.5} $\pm$ \textbf{2.84}\%    & $42.8 \pm 2.83$\%      & $39.8 \pm 0.79$\% & \textbf{41.3} $\pm$ \textbf{0.80}\%  \\
Falcon-7B  
& \textbf{40.2} $\pm$ \textbf{2.81}\%    & $25.9 \pm 2.51$\%      &   \textbf{29.1} $\pm$ \textbf{0.75}\%    & $25.4 \pm 0.72$\%   \\
Mistral-7B 
& $50.1 \pm 2.86$\%    & \textbf{72.4} $\pm$ \textbf{2.56}\%      &   $48.3 \pm 0.80$\%   & \textbf{58.6} $\pm$ \textbf{0.77}\%   \\
Mixtral-8x7B     
& $56.7 \pm 2.84$\%    & \textbf{81.3} $\pm$ \textbf{2.23}\%      & $59.7 \pm 0.77$\%      & \textbf{67.1} $\pm$ \textbf{0.72}\%  \\
\bottomrule
\end{tabular}
\end{table} 

This is especially problematic due to the current trend of naively comparing and transferring numbers across models' technical reports. \citet{ClementineMMLU} demonstrate this concretely: comparing three independent implementations of MMLU (HELM, \evalharness, and the original MMLU code), they find that these implementations produce widely different scores and even change the ranking order of models on the rd---despite ostensibly evaluating the same benchmark.
While they do not report their precise evaluation set-up, the ``MMLU-style'' prompt appears to produce scores consistent with those reported in \citet{jiang2024mixtral} for 25-shot ARC Challenge on Mixtral-8x7B and Llama 2-70B, while Cloze-style or other prompts we tested do not. 
We thus believe Mistral-7B, Mixtral-8x7B, and the Llama models publish results on MMLU but also ARC using this MMLU-style prompt--a decision which is not inherently unreasonable, but which is not documented to the best of our knowledge in technical reports, confusing a variety of practitioners and confounding attempts to replicate the reported scores. 

A further challenge leading to similarly uncertain or inconclusive comparisons is that many headline LLM evaluation results are impossible to reproduce or rerun, because the models themselves have not been made publicly usable. For example, \citet{touvron2023llama1} are forced to pull numbers from the PaLM and Chinchilla technical reports \citep{chowdhery2023pathways,chinchilla} directly for baseline comparison, with no option to rerun these models on Llama's evaluation implementations since PaLM and Chinchilla are not publicly available. 
We do not intend to single out the Llama models as an example: simply to illustrate that many details are either hard or impossible to intuit from publications, especially without context. Indeed, Llama 1 is easier to critique because they do comparatively report more information than average--details such as loglikelihood normalization methodology (see Appendix \ref{app:formal-measurements} for a full description of such details) are documented across the tasks evaluated.

While the prompt is a relatively ``obvious'' location to look when tracking down disparities in reproduced and reported scores, many other equally-important details can also have the same outsized impacts on model performance. For example, \citet{shortformer} demonstrate how the methodology typically used--``nonoverlapping windows'' (Appendix \ref{app:formal-measurements})--to compute perplexity on greater-than-context-length sequences systematically skews results in favor of models trained and thus evaluated with longer context lengths. Awareness of such subtleties' importance has increased, especially for massively long-context models \citep{peng2023yarn}, but it is still underdiscussed relative to its outsized impact, and there is not a clear path for new practitioners to familiarize themselves with such tacitly-known best practices.

\paragraph{Each Task Requires Care and Individual Implementation Considerations}

These supposedly-minor implementation details become even more important when accounting for the need to reimplement every task of interest independently. Most language model releases report scores across a vast range of tasks \citep{OpenAI_GPT4_2023, touvron2023llama1, touvron2023llama2}. Each individual task can be very effort-intensive to implement correctly --- as  \citet{ganguli2023challenges} discuss, even a single evaluation such as BBQ \citep{parrish2021bbq} can take a significant number of researcher-hours to implement well and understand --- which increases the difficulty of vetting all tested tasks for especially smaller teams. \citet{ganguli2023challenges} go on to say that ``Our experience suggests that it is necessary to make many tricky judgment calls and considerations when running such a (supposedly) simple and standardized evaluation.''--this corroborates our experience with~\evalharness. No task is truly trivial to implement and understand with confidence, and often requires discussing with benchmark creators to verify intent.

Further, all benchmarks come with their own subtle quirks and characteristics, which must be handled correctly to avoid skewing results. For example, the AI2 ARC \cite{clark2018think} benchmark is almost entirely 4-choice MCQA, but a single question has 5 possible answer choices instead of 4. MMLU \citep{hendrycks2020measuring} is correctly calculated via performing the \textit{micro} average by averaging accuracy across documents\footnote{\href{https://github.com/hendrycks/test/blob/master/evaluate.py}{As performed by the official MMLU evaluation script }}, rather than \textit{macro}-averaging unweighted accuracies of each of the 57 subjects--the choice of which affects results by several percentage points. This particular detail often is not reported but must be inferred if per-subject MMLU scores are provided. HumanEval \citep{kapoor2024agentevals} has 3 documents lacking the example tests that all other documents possess. We provide these examples not as an exhaustive list, but to show that such subtleties are widespread and affect some of the most popular LM benchmarks.

\paragraph{Crucial Details Go Unreported}

The major reason that such minor implementation choices are so noteworthy is that \textbf{crucial methodological details are insufficiently documented and reported in most works}. Because subtle implementation choices can have an outsized impact on evaluation results, if evaluation code is not published, evaluation setups are likely to be underdetermined based on a publication's description in isolation, through one or more practices being elided from reporting on methodology. For instance, reporting full prompts is a \textit{necessary} but not \textit{sufficient} bar for achieving perfect reproducibility. Despite being insufficient on its own, releasing prompts is vitally important and we heavily recommend it and other methodological disclosures.

It is common practice for researchers to describe evaluation methodology by referring to their use of the setup found in a specific prior work. However, such prior work may not release code, or insufficiently describe its own setup, making comparison difficult. It additionally makes tracing the ``accepted'' practices challenging, especially for newcomers to the field. Many works, for example, trace their methodologies for prompted evaluation to \citet{brown2020language}, who rather thoroughly document the majority of their evaluation setups, despite not releasing code. However, beyond this work, a large number of the foundational approaches used for such few-shot evaluation such as loglikelihood-based multiple choice is not detailed outside of an appendix written by \citet{brown2020language} which does not include several tweaks introduced by subsequent work by other researchers that have become commonplace, making it extremely difficult for newcomers to learn this tacit knowledge about how evaluations are performed. In the hopes of making this folk knowledge more explicit, we document these practices in Appendix \ref{app:formal-measurements}.

\subsection{Best Practices}

In response to these challenges faced in LM benchmarking, we propose several best practices as principles to follow, in order to mitigate the negative impacts.

\begin{itemize}
    \item \textbf{Provide Evaluation Code}: Whenever possible, new publications should be accompanied by \textit{release of the exact evaluation code used.} We acknowledge that this may be difficult, if for example evaluation code is tied tightly to internal software infrastructure\footnote{Though high-effort, code releases such as \href{https://github.com/openai/simple-evals}{https://github.com/openai/simple-evals } demonstrate separation is possible.}--however, code is essential to avoid eliding small, unnoticed yet crucial details, even if the authors may assume these details can be inferred or taken for granted.
    \item \textbf{Provide Prompts and Detail Methodology Thoroughly}: Failing the release of code or to supplement the release of code, prompts should be clearly reported \textit{within a publication.} Other methodological details should be thoroughly described, including, but not limited to: the amount of prompt engineering performed and on which models and evaluation sets; hyperparameters used to calculate loglikelihood or perplexity-based evaluations; generation hyperparameters and answer extraction heuristics used to perform generative evaluations and more. Indirection should be minimized, and at least one source of full ground truth, directly within or linked by the publication, should exist.
    \item \textbf{Create Reporting Standards}: These issues are exacerbated by the fact that there does not exist commonly held practices or checklists within the community to incentivize or ensure adequately thorough documentation of evaluation methodologies. We call upon the community to explore the creation and implementation of such standards.
\end{itemize}

\subsection{Operationalizing Best Practices}

Even slight improvements in the state of evaluation communication, awareness, and reporting can significantly improve evaluation rigor and trustworthiness in the LM research field. Releasing and documenting evaluation code or properly checking task implementations for a wide variety of tasks can be a significant overhead for smaller or lower-resourced teams. By using easier-to-run unified evaluation libraries such as ~\evalharness~or HELM~\citep{liang2023holistic}, one can have a single location for many existing benchmark implementations and remove the need to wade through bespoke codebases and implementations for each novel benchmark. Reporting the use of particular tasks within~\evalharness~on a certain codebase version has become a \textit{de facto} reporting standard for users which allows easier comparisons to other works and makes sharing the code used for one's evaluations easy. 

We have also adapted \evalharness~over its lifetime to date to make it better able to serve the research community. As part of the design for \evalharness, we have made prompt modifications as simple as modifying a single line of a configuration file \footnote{ \href{https://github.com/EleutherAI/lm-evaluation-harness/blob/main/docs/new_task_guide.md\#writing-a-prompt-template}{Writing a Prompt Template in the LM Evaluation Harness }}. This allows prompts to be modified if desired, while otherwise enforcing a sensible default based on community sentiment and common usage for those simply seeking to produce a clean comparison to prior work. Particular task configurations desired by a user can now be communicated through a single configuration file along with a linked library version. Thus, the \evalharness~codebase is not only for reporting common numbers out-of-the-box but also to raise awareness and aiding research studies on the evaluation difficulties we have described \citep{lyu2024probabilities, alzahrani2024benchmarks}.

In general, we following the following priority list for deciding prompting and other evalution details:
\begin{enumerate}
    \item If there is widespread agreement among people who train LLMs, use the agreed upon procedure.
    \item If there is a clear and unambiguous official implementation, use that procedure.
    \item If there is widespread agreement among people who evaluate LLMs, use the agreed upon procedure.
    \item If there are multiple common implementations but not universal or widespread agreement, use our preferred option among the common implementations. As before, prioritize choosing from among the implementations found in LLM training papers.
\end{enumerate}

We choose to prioritize the decisions made by model \textit{trainers} for a few key reasons. First, we have found that people who train models are harder to influence in their practices than people who primarily evaluate models in our work to promote standardization within the ecosystem. Second, many people who train large scale models don't allow others to access their models in the form reported in their reports. Thus people are forced to copy their practices when seeking to have consistent set-ups across models. Third, we have found that as model capabilities have changed (see \Cref{sec:fast-changing}) model developers tend to be more attuned to the most productive ways to evaluate cutting edge models (in part due to having first access to such models). Fourth, people who train models have substantial social and political power within the field and so a large fraction of users will align to their practices regardless.

Turning to validating implementations, we explicitly provide functionality designed around easing qualitative inspection of model outputs and sanity-checking of scores: \texttt{--limit} and \texttt{--log\_samples} flags \footnote{\href{https://github.com/EleutherAI/lm-evaluation-harness/blob/main/docs/interface.md\#command-line-interface}{LM Evaluation Harness Command Line Interface }} can be used to dry-run evaluations on a small subset of samples for testing and prototyping, and per-sample scores can be recorded and logged to disk, while also saving model outputs for post-hoc reproducibility without additional compute expenditures.

Although more work should be done to improve reporting standards around evaluation, we believe that a common research infrastructure can significantly ease the burden of good practices for researchers, reduce duplicated effort massively, and increase confidence in evaluation scores that use such common libraries by and large.

\section{Evaluations are Not Just Numbers}\label{sec:Evaluations are Not Just Numbers}

When one runs an evaluation benchmark, what comes out is a numerical score. Those numbers are often \textit{used by people} to assess the relative capabilities of models, determine their suitability for deployment, or consider the need for specific rules and regulations. The appropriate conclusions to draw about a model's behavior are \textit{informed by}, not \textit{constituted by}, these numerical scores. There is currently a substantial lack of attention to this fact, and replacing genuine evaluation with computing benchmark scores is substantially hindering language model evaluation, research, and governance.

\subsection{Challenges}

\paragraph{What, if Anything, Does a Benchmark Measure?} It is almost never the case that one cares about the \textit{actual score} of a language model on a particular benchmark. Instead, benchmarks are typically proxies for some broader notion of capability or ability that is difficult to make precise and measure. While being able to rapidly evaluate models and compare their capabilities numerically is very useful in many contexts, overemphasis and focus on particular numeric scores can hinder progress or prevent accurate senses of performance. This is especially true as precise numerical scores are very sensitive to evaluation settings.

While debate around the validity of machine learning benchmarks has been widespread for years \citep{messick1994validity,raji2021ai,saphra-tragedy-2023,davis2023benchmarks,subramonian2023takes}, 
it has largely been ignored by those developing the latest wave of language models and language model evaluation benchmarks. We find this particularly concerning in the context of evaluations informing deployment decisions: it is unclear (or in some cases very clear to the contrary) that evaluations, especially more artificially-constructed benchmarks such as minimal-pair or pronoun-frequency \citep{nangia-etal-2020-crows, zhao2018gender, neveol:hal-03629677} based bias benchmarks or multiple-choice knowledge tests \citep{hendrycks2020measuring}, provide information about how models will perform in more realistic deployment scenarios. Indeed, research like \citet{lyu2024probabilities} demonstrate that multiple-choice loglikelihood-based benchmarks do not even agree with generative-scored multiple-choice benchmarks. \citet{omiye2023large} and \citet{hofmann2024dialect} find that current safety-trained models, despite exhibiting apparently-safer or less bias on current bias benchmarks, amplify biases far beyond what is suggested by their training data when engaged in practical, real-world interactions such as in high-risk settings like medicine.


\paragraph{Lack of Reporting on Uncertainty and Variance}  Large language models are heavily stochastic objects. Their behavior depends on a large number of both arbitrary decisions and random events that occur during their training and evaluation. This fact is very rarely acknowledged during model evaluation, likely because it makes evaluating models much more challenging and requires statistical expertise most machine learning researchers lack. However, omitting such information can paint highly misleading pictures of model performance. This issue is becoming more acute over time--evaluation benchmarks are increasingly focusing on advanced concepts or capabilities that can be challenging to produce examples for \textit{en masse}. This results in smaller evaluation datasets, and correspondingly larger uncertainty in numeric scores. 

In an excellent blog post on this topic, \citet{spend-more} presents the following two pictures of GPQA performance for some of OpenAI and Anthropic's models:

\begin{figure}[h]
    \centering
    \includegraphics[width=\linewidth]{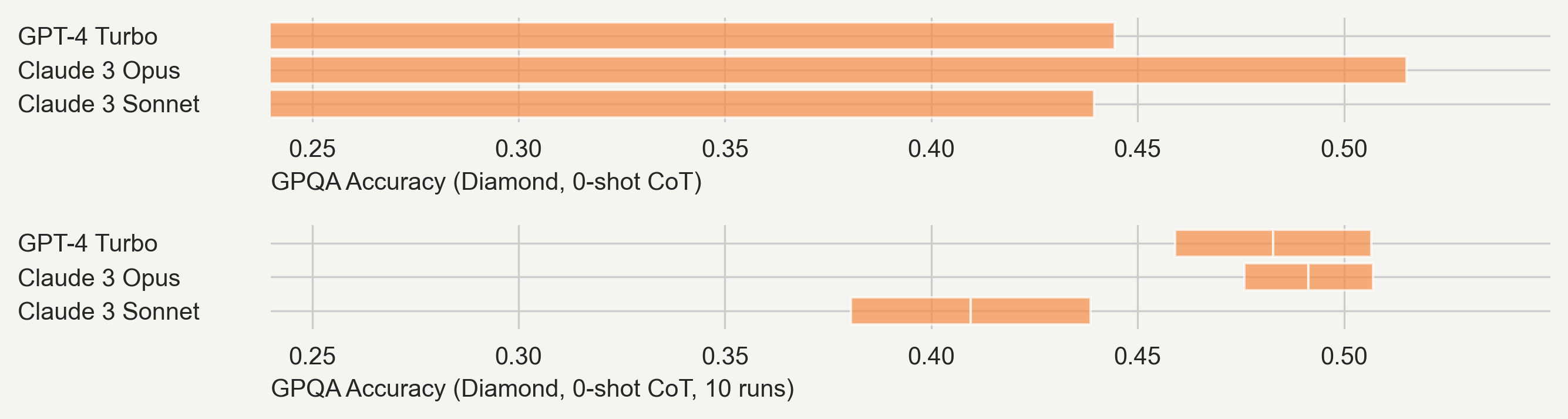}
    \caption{Plots showing GPQA performance of various models in a single run (top) and a confidence interval computed after doing ten runs (bottom). The two plots lead to very different conclusions about the relative performance of models. Data from \citet{spend-more}, presentation edited.}
    \label{fig:gpqa}
\end{figure}

Presenting single-run point estimates of performance (top) instead of a 95\% confidence interval across 10 runs of the benchmark (bottom) paints a very different pictures regarding the comparative performance of different models. Similar behavior holds for benchmarks such as HumanEval \citep{chen2021codex}, which contains only 164 examples. Practitioners frequently tout small increases in numeric score on datasets such as these, ignoring the fact that such ``improvements'' might be washed out simply via sampling again at the same temperature.

\paragraph{The Benchmark Lottery} The Benchmark Lottery \citep{dehghani2021benchmark} posits that popular benchmarks can have a distorting effect on the field, redefining success in ways that reinforce their particular idiosyncrasies and even flaws. A concrete illustration of this phenomenon can be seen in the Open LLM Leaderboard \citep{open-llm-leaderboard}, where the choice of evaluation format has itself become a target for optimization: practitioners have been observed fine-tuning their models specifically to match the evaluation format used by the leaderboard, such as targeting separator tokens used in particular versions of~\evalharness, rather than improving the underlying capabilities being measured. The leaderboard's arbitrary choices about which benchmarks to include and which implementation details to adopt effectively define what ``good'' means for a large segment of the open-source community, making it a high-stakes example of how benchmark selection and implementation choices can distort incentives.

The sensitivity to implementation choices underlying this phenomenon is also illustrated in \cref{tab:prompt-sensitivity}. We show that GPT-NeoX-20B performs approximately randomly on both ARC-C and MMLU using what we term MMLU-style evaluation, but much better using other prompting formats. Through exploratory analysis of model completions we have observed that this at-random performance is a result of failing to follow the task and output format for this prompting style, even when it can score well using an alternative format. An even more extreme example is from ARC-Easy: using the ``cloze'' evaluation setup, GPT-NeoX achieves an accuracy of $72.4 \pm 1.80\%$ but using the MMLU-style setup it scores $26.5 \pm 1.78\%$.

While the Open LLM Leaderboard example shows the impact of explicit gaming, these results show a phenomenon that is much harder to guard against: in a different but plausible world in which MMLU-style prompting was the standard introduced in \citet{brown2020language}, the perceived performance of GPT-NeoX-20B and other models trained on the Pile \citep{gao2020pile} would be extremely low, despite strong results that would be observed in the ``cloze'' setting. In such a world, we very plausibly could have concluded that our dataset was garbage (instead of concluding that it didn't enable performance in a specific format only) and wasted many human and GPU-hours trying to debug the wrong problem.

These issues are so extreme that it is common for people who work with language models to discuss ``vibes tests,'' the intuitive feeling they get about a model after using it for an hour or so, as being as informative or even more informative than the benchmarks that much of our field is allegedly based on. This sentiment is widespread enough that recent work has sought to formalize it: \citet{dunlap2024vibecheck} introduce a framework for discovering and quantifying the qualitative differences between models that users notice through informal interaction but that standard benchmarks fail to capture, while Reka AI's Vibe-Eval \citep{padlewski2024vibeeval} explicitly designs evaluation suites around these hard-to-quantify aspects of model quality. The Chatbot Arena \citep{DBLP:journals/corr/abs-2306-05685} can be viewed in a similar light---as an effort to scale human preference judgments into a systematic ranking methodology, producing orderings that frequently diverge from those derived from standard benchmarks. That the community has found it necessary to build evaluation methodology around formalized intuition is itself a powerful indictment of the current state of benchmarking.

\subsection{Best Practices}

We propose several recommendations to follow to improve the state of LM evaluation on this front:

\begin{itemize}
    \item \textbf{Perform Statistical Analyses, and Report on Sources of Variance and Error}: Instead of reporting isolated numbers, more effort should be made to contextualize the durability of performance estimates made by benchmarks, such as by reporting variance over several runs, or reporting statistical significance. These practices are seldom done in the field.
    \item \textbf{Evaluate Models in Realistic Settings}: Popular knowledge-centric evaluation benchmarks use prompting set-ups that are much more similar to how examination questions for humans are written than how people actually use language models \citep{son2024kmmlu,rein2023gpqa}. To improve the usefulness and validity of evaluations, more effort should be made to create test environments and datasets that mirror the real-world use-cases and deployment settings desired for a given model. 
    \item \textbf{Do Exploratory Analysis to Build Better Understanding of Results}: To understand why a model is scoring so well or so poorly, it is important to do some sort of qualitative error analysis. This can sometimes reveal superficial errors that are easier to correct with post-processing~\citep{ClementineDROP, bawden-yvon-bloom-mt-2023}, or more fundamental errors. Such practices allow more holistic conclusions and comparisons between methods, and increase confidence in evaluation correctness. 
\end{itemize}

\subsection{Operationalizing Best Practices}

Improving the interpretation and nuanced consideration of benchmarks is primarily an issue of \textit{effort} and \textit{defaults}. Because considerations such as reporting statistical significance or measures of uncertainty require additional effort beyond the ``standard'' of simply receiving a higher-is-better evaluation number to report, they are not commonly done. We have sought to shift this status quo, as with our goals around reproducibility, via by-default opting users of~\evalharness~ into better evaluation reporting. We report bootstrapped standard error metrics by default, making reporting of confidence intervals (as we have done in this paper) nearly as simple as copy-pasting an additional number into one's paper.

It is important to note that bootstrap confidence intervals are a slightly different notion of variation than the example given in Section 4.1. Language models are stochastic objects, so one natural notion of variation would be to run the model several times and measure how the results change (as done in Section 4.1). The bootstrap CIs offered in~\evalharness~ by contrast are oriented towards answering a different question: suppose that the particular questions in a given evaluation benchmark are samples from some distribution of interesting questions in the world. How much variation in model behavior should we expect \textit{as we sample new questions from the same distribution}? We believe that both of these notions of variation (as well as variation across prompts and scoring systems, as previously discussed) are interesting and valuable.

We also seek to encourage future study of variation across prompts--as part of the
BigScience Research Workshop\footnote{\href{https://bigscience.huggingface.co}{https://bigscience.huggingface.co}}, we developed tooling using the PromptSource library \citep{bach2022promptsource} for \textit{multiprompt evaluation}, enabling users to run many different prompts for the same task and analyze the variance across prompt choices. Most papers that came out of the BigScience Workshop used this functionality to report distributions of scores across different prompting set-ups \citep{sanh2022multitask,muennighoff2022crosslingual,yong2023bloom,workshop2023bloom}. PromptSource, along with other innovations introduced in the BigScience fork\footnote{\href{https://github.com/bigscience-workshop/lm-evaluation-harness}{https://github.com/bigscience-workshop/lm-evaluation-harness}}, is now supported natively in \evalharness. 


\section{Conclusion}

In this manuscript, we have highlighted a number of challenges that are ubiquitous in LM evaluation. We have highlighted and attempted to document a number of items of ``folk knowledge'' around benchmarking best practices that is either barely-documented or scattered across many sources. We then have highlighted our recommendations for mitigating many of these challenges' impacts, and discussed how, in our work, we have incorporated such practices into the design of~\evalharness, which has been widely used to mitigate the impact of such evaluation challenges for many practitioners. We hope that this work will draw more attention to the shortcomings of current LM evaluation, and will serve as a useful resource to practitioners in performing rigorous evaluation and charting effective paths forward.

\bibliography{bibliography}
\bibliographystyle{acl_natbib}

\section*{Checklist}

\begin{enumerate}

\item For all authors...
\begin{enumerate}
  \item Do the main claims made in the abstract and introduction accurately reflect the paper's contributions and scope?
    \answerYes{}
  \item Did you describe the limitations of your work?
    \answerYes{} See Appendix \ref{app:limitations} for a brief standalone discussion of limitations. We additionally discuss our scope and are honest about our limitations of our work in the main text throughout.
  \item Did you discuss any potential negative societal impacts of your work?
    \answerYes{} See Appendix \ref{app:impacts} for a discussion of potential societal impacts.
  \item Have you read the ethics review guidelines and ensured that your paper conforms to them?
    \answerYes{}
\end{enumerate}

\item If you are including theoretical results...
\begin{enumerate}
  \item Did you state the full set of assumptions of all theoretical results?
    \answerNA{}
	\item Did you include complete proofs of all theoretical results?
    \answerNA{}
\end{enumerate}

\item If you ran experiments (e.g. for benchmarks)...
\begin{enumerate}
  \item Did you include the code, data, and instructions needed to reproduce the main experimental results (either in the supplemental material or as a URL)?
    \answerYes{} 
  \item Did you specify all the training details (e.g., data splits, hyperparameters, how they were chosen)?
    \answerYes{}
	\item Did you report error bars (e.g., with respect to the random seed after running experiments multiple times)?
    \answerYes{} Yes, for all experiments we perform as case studies, we report 95\% confidence intervals calculated via bootstrapping.~\evalharness makes reporting such numbers trivial.
	\item Did you include the total amount of compute and the type of resources used (e.g., type of GPUs, internal cluster, or cloud provider)?
    \answerYes{}
\end{enumerate}

\item If you are using existing assets (e.g., code, data, models) or curating/releasing new assets...
\begin{enumerate}
  \item If your work uses existing assets, did you cite the creators?
    \answerYes{} We cite all directly referenced benchmark datasets, and~ \evalharness provides citations to all implemented tasks that we do not use or reference here.
  \item Did you mention the license of the assets?
    \answerNA{}
  \item Did you include any new assets either in the supplemental material or as a URL?
    \answerNA{}
  \item Did you discuss whether and how consent was obtained from people whose data you're using/curating?
    \answerNA{}
  \item Did you discuss whether the data you are using/curating contains personally identifiable information or offensive content?
    \answerNA{}
\end{enumerate}

\item If you used crowdsourcing or conducted research with human subjects...
\begin{enumerate}
  \item Did you include the full text of instructions given to participants and screenshots, if applicable?
    \answerNA{}
  \item Did you describe any potential participant risks, with links to Institutional Review Board (IRB) approvals, if applicable?
    \answerNA{}
  \item Did you include the estimated hourly wage paid to participants and the total amount spent on participant compensation?
    \answerNA{}
\end{enumerate}

\end{enumerate}


\appendix

\section{Library Design}\label{app:library-design}

Here we provide an overview of the major components and design philosophy of~\evalharness. At its core,~\evalharness~allows for the contribution of two types of implementations: evaluation \textbf{Tasks} and integrations with novel \textbf{LM} implementations.

\paragraph{Tasks}\label{app:evalharness-tasks}

\begin{figure}[h]
    \centering
    \includegraphics[width=\linewidth]{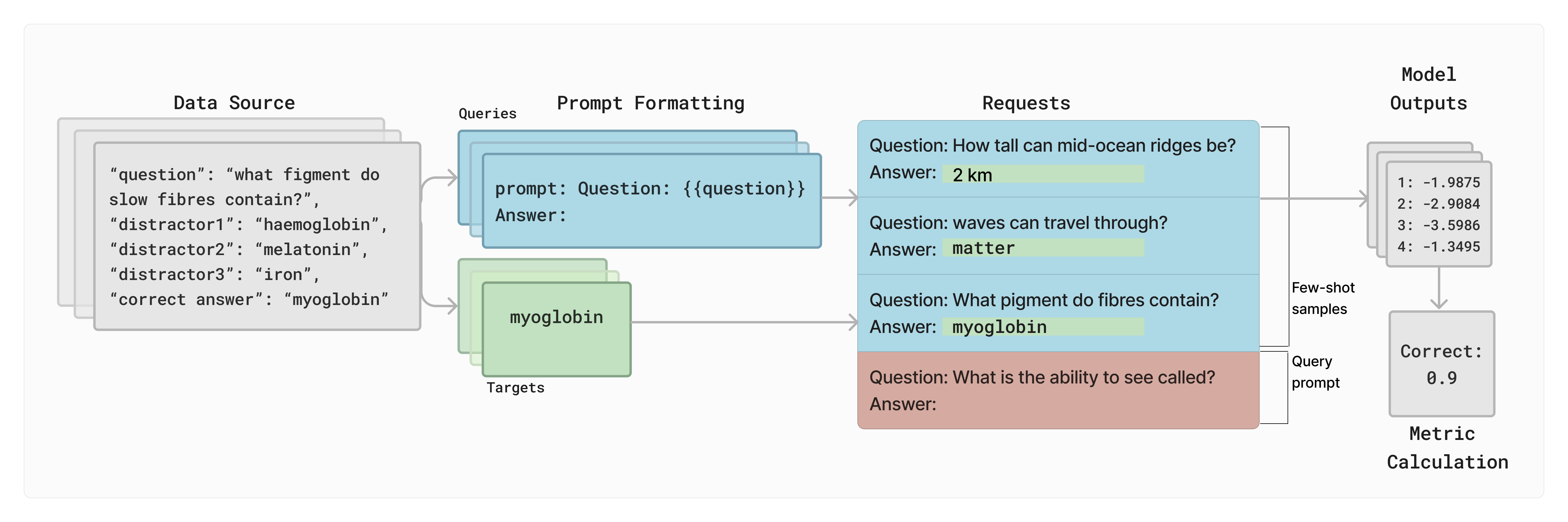}
    \caption{The operations performed by a \textbf{Task} object in \evalharness. Tasks are configured by YAML files or as a Python subclass, and encompass 1) a data source (using the \texttt{Datasets} library~\citep{lhoest-etal-2021-datasets}), 2) tools for defining prompts and format, 3) mapping these prompts to rendered inputs and expected output type from an \textbf{LM} in the form of \textbf{Requests}, and 4) rules for post-processing the \textbf{LM}'s outputs and calculating the final task metrics.}
    \label{fig:task-object}
\end{figure}

\evalharness~is built around modular implementations of evaluation tasks, implemented as a \textbf{Task} class using a common API. This allows tasks to be collected in a common library, for new tasks to be extended or implemented easily, and for novel tasks to be easily shared reproducibly among practitioners or other library users. Users can implement tasks either via YAML-based configuration files or via subclassing the provided \textbf{Task} class and providing custom code for specific methods. In Figure \ref{fig:task-object}, we show an example of the evaluation logic packaged within a \textbf{Task} class.

We provide a number of implementations for common tasks, and accept new tasks sourced from the community. We strive to match the paper originally introducing a benchmark dataset in its methodology, including using the same prompts if applicable. For tasks such as those introduced prior to prompted evaluation becoming the standard, we source evaluation methodology from the paper first posing the evaluation dataset as a prompted task. For example, we implement many tasks as adapted for in-context learning by \citet{brown2020language}.

\paragraph{LMs} The next core piece of~\evalharness~is the \textbf{LM} API. Because effective \textit{orchestration} is our core goal, we allow arbitrary software libraries or (autoregressive) language model architectures to extend a provided interface for \textbf{LM} objects.

For ease of use, and compartmentalization of the model definition and external library integrations for custom models away from core evaluation logic, we assume that LMs operate upon dispatched \textbf{Requests} which consist of mapping \textit{string inputs} to some \textit{string or probability} as output. We thus abstract tokenizers away within the \textbf{LM} class, and treat a neural language model combined with its tokenizer as a single system being evaluated.

LMs implement a simple interface, consisting of several types of \textbf{Requests} in order to be used within the library for all supported tasks.

\paragraph{Request Types} We allow for 3 core types of \textbf{Requests} that may be sent to a language model, which consist of distinct types of \textit{measurements} that can be performed to observe a model's response or latent capabilities in a prompted format. These are:

\begin{itemize}
    \item (Conditional) Loglikelihoods (\texttt{loglikelihood, multiple\_choice}) - computing the probability of given output string(s), conditioned on some provided input.
    \item Perplexities (\texttt{loglikelihood\_rolling}) - measuring the average loglikelihood or probability of producing the tokens in a given dataset.
    \item Generation (\texttt{generate\_until}) - generating text until a given stopping condition is reached, from a model conditioned on some provided input.
\end{itemize}

\begin{figure}[h]
    \centering
    \includegraphics[width=\linewidth]{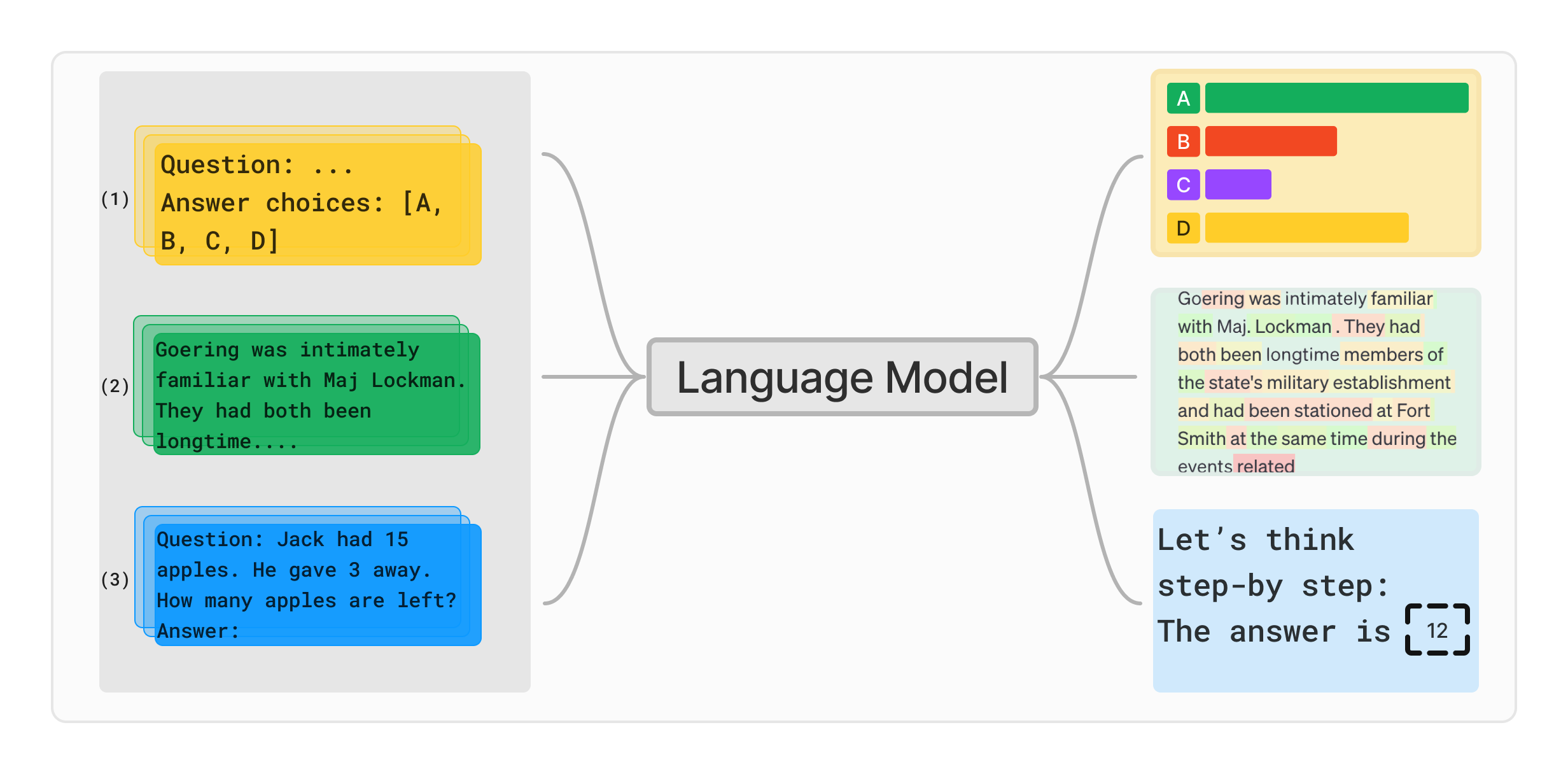}
    \caption{Overview of the three core \textbf{Request} types supported by our evaluation framework. These include (1) conditional loglikelihoods, (2) perplexities, and (3) generation-based requests.}
    \label{fig:task-types}
\end{figure}

Provided with these three primitive operations, we are able to implement the major ways in the literature that have been used to evaluate LMs (\citet{gao2020pile}, \citet{brown2020language}, \textit{inter alia}). While these high-level approaches are standard, they all contain a number of subtle implementation decisions which are often not disclosed in papers. We include a full formal description of common implementation details involved in ours and others' approaches in Appendix \ref{app:formal-measurements}.

\section{Formalizing Measurements}\label{app:formal-measurements}

Here we provide a formal description of the most common approaches to obtaining outputs or measurements from LMs for evaluation, as we implement in \evalharness . We include this for a number of reasons: as a reference for future work, notes on the history of certain LM eval practices, and as an illustrative example of just how many implementation details or methodological choices \textit{do not} typically make it into evaluation papers and yet can vitally impact results or findings.

\subsection{Preliminaries}

Throughout, we consider an auto-regressive language model (LM), with vocabulary $V$. Given an input consisting of tokens $x_0, x_1, ..., x_{n-1}$, the model outputs a probability distribution over the vocabulary, $P(x_{n} | x_0,x_1,...,x_{n-1})$. Internally, this is represented as returning ``logits'' of shape $(1, |V|)$, which when taking a log-softmax over the vocabulary dimension, yields \textit{log probabilities} (``logprobs'' or ``loglikelihoods'') of each token in $V$. Logits are the raw, unnormalized predictions of the model before applying the softmax function. Crucially, due to the parallel training and causal masking of autoregressive LMs, it is possible to obtain from a single LM call with $x_0, x_1, ..., x_{n-1}$ as input, logits of shape $(n, |V|)$ with the $i$-th element of these logits representing $P(x_{i} | x_0,x_1,...,x_{i-1})$ for all $1 \leq i \leq n$. (That is, for every token position of the input, we obtain concurrently the model's prediction for the subsequent token, starting from its prediction for the value of $x_1$ and ending with the model's predictions for the (not provided) ``$x_n$'' token.)

\subsection{Ranking-Based Multiple Choice QA}\label{app:mcqa-cond-loglikelihood}

Given our language model, we aim to compute the conditional (log) probability (or ``loglikelihood'') of a target string $y$ conditioned on input $x$, denoted as $\log P(y|x)$. This can be performed in a single LM call.

Let $x = x_0, x_1, ..., x_{n-1}$ be an input sequence of $n$ tokens and $y = y_{0}, y_{1}, ..., y_{m-1}$ be the target sequence of $m$ tokens, where $x_i$ and $y_i$ represent individual tokens. To compute $\log P(y|x)$, we follow these steps:

\begin{enumerate}
    \item Concatenate $x$ and $y$ to form a new sequence, but discard the final token $y_{m-1}$. The resulting sequence is $x_0, x_1, ..., x_{n-1}, y_{0}, y_{1}, ..., y_{m-2}$.
    \item Pass this concatenated sequence through the language model to obtain logits $l$ of shape $(n + m - 1, |V|)$, where $|V|$ is the size of the vocabulary. The last $m$ positions in these logits correspond to the predicted probability distributions for the target tokens $y_0$ to $y_{m-1}$, conditioned on the input $x$ and the preceding target tokens.
    \item Apply a log-softmax function to the last $m$ logits to obtain log probabilities for the completion tokens only.
    \item Calculate the conditional loglikelihood of the target string $y$ given the input $x$ by summing the log probabilities of each target token:
    \begin{equation} \label{eqn:logp}
    \log P(y|x) = \sum_{i=0}^{m-1} \log p(y_{i}|x, y_0, ..., y_{i-1}) = \sum_{i=0}^{m-1} l(n +i, y_i),
    \end{equation} where $\log p(y_i|x, y_0, ..., y_{i-1})$ is the log probability of the $i$-th target token conditioned on the full input $x$ and the preceding target tokens. (and where $x, y_0,... y_{-1}$ denotes conditioning on only $x$.)
\end{enumerate}

With this primitive for computing $\log P(y|x)$, several options for evaluation (and decisions regarding hyperparameters) become available.

\textbf{Multiple Choice QA}

Equation \ref{eqn:logp} determines how to compute $\log P(y|x)$. We now describe how to perform \textit{loglikelihood-based multiple choice} as described by \citet{brown2020language}: given $k$ possible answer strings $a_1, a_2, ..., a_k$, we compute the model's answer to be $\texttt{argmax}(\log P(a_1|x), \log P(a_2|x), ..., \log P(a_k|x))$. In other words, the model selects the answer string with the highest conditional log probability given the input $x$.

This can be performed with worst-case $k$ LM calls using the approach to calculate $\log P(y|x)$ for each $a_i = y$ described above. However, the number of LM calls can be reduced if one or more answer strings are only a single token in length. Assume some $a_i$ is only encoded by a single token $z$. Then, when calculating the loglikelihood of another answer string $a_0$, we obtain the (log-softmaxed) logits of shape $(n + m - 1, |V|)$ as an intermediate output. These logits contain the predicted log probabilities for each token in the vocabulary at each position, conditioned on the input $x$ and the preceding tokens. To extract the loglikelihood of predicting the single-token answer $a_i$ conditioned on $x$, we can simply select the element in $l$ corresponding to token $z$ at position $n$. This logit represents the log probability of predicting token $z$ immediately after the input sequence $x_0, x_1, ..., x_{n-1}$.

Thus, we can calculate the loglikelihood of a single-token continuation ``for free'' and remove an additional LM call for each such single-token $a_i$.

\textbf{Normalization}

While the above approach uses the raw loglikelihoods of each given answer choice to select a model answer, other options are available. For instance, if each answer string $a_i$ is different in length, this process may frequently default to selecting the shortest $a_i$ simply because loglikelihoods are the sum over individual tokens' log probabilities. Several options for \textit{normalizing} these loglikelihoods are possible, as also described in \citet{mcqa-normalization-eleutherai}:
\begin{itemize}
    \item \textbf{Token-length normalization}: each $a_i$'s loglikelihood is divided by $m_i$, its length in tokens, to gain the per-token loglikelihood of each answer. This approach requires no additional LM calls, and is used alternately with raw loglikelihoods for most tasks by \citet{brown2020language}.
    \item \textbf{Byte-length normalization}: each $a_i$'s loglikelihood is divided by its length in bytes, removing the dependence on the model's tokenizer but still normalizing by answer string length.~\evalharness~provides this metric where applicable as \texttt{acc\_norm}.
    \item \textbf{Mutual Information}: each $a_i$'s loglikelihood is defined as $\log P(a_i|x) - \log P(a_i|null)$, where $null$ is either the empty string, a BOS token, or a placeholder such as \texttt{"Answer:"}. This can be thought of as a notion of the \textit{pointwise mutual information} \citep{shannon1948, askell2021general}, $\log\left(\frac{P(a_i|x)}{P(a_i)}\right)$, which measures the increase in the likelihood of outputting $a_i$ when conditioned on the input $x$, compared to the likelihood of outputting $a_i$ unconditionally. Intuitively, this measure of mutual information captures the extent to which introducing $x$ makes $a_i$ more likely. Although this approach is nonstandard, it is provided in ~\evalharness~ under the option \texttt{acc\_mutual\_info}, and used selectively by \citet{brown2020language} and \citet{askell2021general} for certain tasks.
\end{itemize}


\textbf{Computing Exact Match}

In addition to computing loglikelihoods and normalized loglikelihoods, we may also want to determine whether a given target string $y$ would be produced by greedily decoding from the input $x$. Let $z$ be the concatenation of $x$ and $y$ as defined in the previous sections, and let $\ell$ be the logits 
 of shape $(n + m - 1, |V|)$ obtained by passing $z$ through the language model. 
To compute the exact match, we compute 
$\sum_{i=0}^{m-1} \mathbbm{1}[y_{i} = \argmax(\ell(n + i, \cdots))]$, where $\mathbbm{1}[\cdots]$ is the indicator function that returns 1 if the condition is true and 0 otherwise, and $\ell(n + i, \cdots)$ represents the logits vector corresponding to the model's output logits predicting the $n+1+i$-th token and therefore the $i$-th token position in $y$ (0-indexed). Intuitively, this sum checks whether each token $y_{i}$ in the target string $y$ matches the most probable (argmax) token predicted by the model at each step of greedy decoding. If the sum equals $m$ (the length of $y$), it means that all tokens in $y$ would be produced by greedily generating $m$ tokens starting from $x$. In this case, we return True to indicate an exact match. Otherwise, if the sum is less than $m$, we return False, indicating that $y$ would not be produced by greedy decoding. Computing the exact match can be useful in scenarios where we want to assess whether the model can generate a specific target string verbatim.

\textbf{Tokenization}

In the above derivations, we assume that one can safely tokenize $x$ and $y$ separately and concatenate their tokenizations. This assumption is not always valid, and most widely-used language model tokenizers provide no such guarantees.

While this factor does not impact the validity of the above calculations, we note that this implies one should be very careful with how their prompt will be tokenized, especially in cases where the tokenization of an input and output pair separately may not align with the tokenization the LM most often saw during training. There are some recently proposed mitigations to remedy this issue, such as ``token healing''\footnote{\url{https://github.com/guidance-ai/guidance/blob/main/notebooks/tutorials/token_healing.ipynb}}, 

In the case of \evalharness, we achieve a majority of benefits via shifting trailing prompt whitespace into each target string $y$ \footnote{\url{https://github.com/meta-llama/llama/issues/217\#issuecomment-1774147331}}, and do not include trailing input whitespace for all tasks we implement, so this operation should be null for the vast majority of cases.

We hope that future work can examine the most practical way to remove such tokenization-based concerns and difficulties from evaluation, such as BPE dropout \citep{provilkov2020bpedropout}, other regularization techniques, or other novel tokenization innovations.

\subsection{Perplexity evaluation}

A common approach to measure language modeling performance on some data distribution $D$ is to measure \textit{perplexity}, which is defined as the exponential of the average negative loglikelihood per token \citep{perplexityjelinek, brown1992estimate}, that is:
\begin{equation}
PPL = \exp\left(\frac{-1}{\sum_{j=1}^{|D|}N_j}\sum_{j=1}^{|D|}\sum_{i=1}^{N_j}\log P(y_{j_i}|y_{j_1}, \dots, y_{j_{i-1}})\right),
\end{equation}\label{eqn:ppl} where $|D|$ is the number of documents in the dataset, $y_j$ is the $j$-th document in $D$, $N_j$ is the total number of tokens in $y_j$, and $y_{j_i}$ represents the $i$-th token of $y_j$.

To calculate perplexity on a selected dataset $D$, each dataset document $y$ is tokenized and fed into a language model following the procedure to calculate $\log P(y)$ described in Appendix \ref{app:mcqa-cond-loglikelihood}, via computing $\log P(y|x)$, where $x$ is set to either the empty string or a beginning-of-text token. Thus, given $\log P(y)$, for each document $y \in D$ we can sum up the per-document loglikelihoods and divide by the number of total dataset tokens.  However, comparing perplexity across models that use different tokenizers can be challenging, as the number of tokens per document and the average next-token prediction difficulty will vary. 


\textbf{Tokenization}

To avoid introducing a dependence on tokenizers while reporting perplexity scores, several options are available: 

\begin{itemize}
    \item \textbf{Bits per Byte:}  This metric measures the average number of bits required to encode each byte of the input text, providing a tokenization-agnostic measure of language modeling performance \citep{gao2020pile}. Formally: 
    \begin{equation}
BPB = \frac{-1}{log(2)}\left(\frac{-1}{\sum_{j=1}^{|D|}B_j}\sum_{j=1}^{|D|}\sum_{i=1}^{N_j}\log P(y_{j_i}|y_{j_1}, \dots, y_{j_{i-1}})\right),
\end{equation}\label{eqn:bpb}where $\log$ is in base $e$ and $B_j$ is the length in bytes of document $y_j$. Alternately, bits per byte can be written as
\begin{equation} \label{eqn:bpb2}
BPB = \frac{\sum_{j=1}^{|D|}N_j}{\sum_{j=1}^{|D|}B_j}\log_2(PPL) = \frac{\sum_{j=1}^{|D|}N_j}{\sum_{j=1}^{|D|}B_j}\frac{\log(PPL)}{\log(2)}.
\end{equation}
    That is, taking the base-2 log of perplexity and renormalizing by the number of bytes rather than tokens.
    \item \textbf{Word-Level Perplexity:} By tokenizing the input text into words, such as via splitting on whitespace, we can calculate perplexity based on the average loglikelihood per \textit{word} rather than per-token, making the metric comparable across models with different subword tokenizers.
    \item \textbf{Byte-level Perplexity:} Similarly, calculating perplexity averaged over the number of \textit{bytes} instead allows for a different tokenization-independent perplexity calculation, as the number of bytes in each document's string remains constant regardless of the tokenizer used. 
\end{itemize}

Both byte- and word-level perplexities can be calculated via replacing $N_j$ in Equation \ref{eqn:ppl} instead with the number of bytes or ``words'' in document $j$.

In ~\evalharness~we implement and report all 3 of the above metrics and report them as tokenization-agnostic measures of perplexity. This approach aligns with the work of \citet{gao2020pile}, who popularized the use of bits per byte for measuring perplexity, and has been adopted in subsequent studies such as \citet{palomamagnusson2023} and \citet{chinchilla}.


\textbf{Sliding Window}

Another challenge is the approach taken to measure perplexity on documents longer than the context length of a given LM. A natural approach, as used by \citet{gao2020pile}, is to chunk documents longer than a model's training context size $L$ into non-overlapping chunks. For example, a document of length 4500 tokens evaluated using a model with context length 2048 would be processed as follows: tokens 0:2047 (with token 0 being a prepended BOS token) are fed to predict tokens 1:2048, then tokens 2048:4095 are fed to predict tokens 2049:4096, and finally tokens 4096:4499 are fed to predict tokens 4097:4500. The loglikelihoods of each chunk are then summed to obtain the entire document's loglikelihood. 


However, \citet{shortformer} observe a phenomenon they call the ``Early Token Curse'', referring to the fact that tokens with a greater amount of context preceding them are fundamentally easier to predict, whereas the first several tokens a model must predict ``from scratch'' are difficult or impossible to predict without information to condition on. To mitigate this issue, they propose a \textit{strided} or \textit{sliding window} perplexity evaluation method.

Instead of creating non-overlapping windows of tokens of size $L$, the strided approach introduces a stride $s$ such that overlapping windows of size $L$, shifting at each time by $s$ positions, are used to score $s$ new tokens' loglikelihoods. This is equivalent to \citet{gao2020pile}'s approach when $s = L$. This approach reduces the skew of perplexity favoring models with larger $L$ (and thus fewer tokens appearing at the beginning of a context window) that is introduced by the early token curse via reducing the number of tokens appearing with little context preceding them. However, it is worth noting that the prevalence of such affected tokens decreases with larger context window sizes $L$ for a model.

A downside of the above method is that a naive implementation requires in the worst case, $\frac{L}{s}$ times the calls to an LM and $\frac{L}{s}$ the compute compared to the non-overlapping window approach. (However, some architectures can leverage KV cache reuse to avoid the cost of repeatedly re-encoding tokens). \evalharness~follows \citep{gao2020pile} in using non-overlapping windows of size $L$. We believe this choice balances the computational cost and the mitigation of the early token curse, while still providing a standardized and comparable measure of language modeling performance.

\subsection{Generative Evaluation}

While loglikelihood-based tasks, such as multiple-choice question answering, provide a valuable measure of a language model's understanding and ability to rank given options, they do not directly assess the model's capacity to generate coherent and relevant text. Generative tasks, on the other hand, require the model to produce original text based on the given context.

Generative tasks have gained significant importance in recent times, particularly due to the fact that many popular language model APIs either do not provide\footnote{\url{https://docs.anthropic.com/claude/reference/complete_post}} or greatly limit access\footnote{\url{https://platform.openai.com/docs/api-reference/chat}} to log probabilities or other intrinsic measures of the model's confidence in its outputs. This shift has made it especially necessary to rely on the generated text itself to evaluate the model's performance and capabilities.

\textbf{Sampling Hyperparameters}

Generative tasks often involve various techniques for controlling the diversity and quality of the generated text, such as sampling with temperature, top-k or top-p (nucleus) sampling \citep{nucleasholtzman2020curious}, and beam search \citep{beamsearchli-etal-2016-deep}. The choice of these hyperparameters can significantly impact the model's output and, consequently, its performance on the task. It is essential to consider and report these hyperparameters when evaluating generative models, as they can greatly influence the generated text's characteristics and the model's overall performance.

\textbf{Scoring and Answer Extraction}

Due to the challenges in measuring language model output (Section \ref{sec:background}), particularly in verifying the semantics of natural language, and because free-form generation sacrifices the benefit of the artificially restricted input space of multiple-choice tasks, the challenge of scoring answers for quality or correctness must be tackled differently.

The open-ended nature of generative tasks means that there may be multiple valid and appropriate responses to a given prompt. A common evaluation strategy is to use few-shot prompts, where the model is provided with a number of examples demonstrating the desired input-output format. The model is then prompted with a new input, and its generated response is extracted using regular expressions (regex) or other heuristic approaches to obtain the normalized answer strings which can be evaluated using exact-match or other metrics. This approach allows for a more structured evaluation of the model's ability to generate accurate and relevant responses based on the given examples.

However, this is often a highly imperfect solution, as different models may generate responses in varying formats, making it challenging to create a universal regex pattern that works for all models. Moreover, the effectiveness of the regex-based extraction is highly dependent on the specific format used in the original task creation, which could introduce bias towards models that generate responses in a similar format. To address these limitations, \evalharness~provides a highly customized answer extraction mechanism through a \textbf{Filter} component tied to \textbf{Task} implementations, allowing the model output to be put through an arbitrary number of filters and post-processing steps.

These custom heuristic approaches make the release of evaluation code, and our recommendations in general, even more crucial. Without knowledge of the extent to what extraction code is used, how it may be tailored to a model, or without access to model outputs, it is difficult to separate models' compliance with the evaluation \textit{format} from their answer \textit{correctness}. Therefore, it is essential to provide detailed information about the answer extraction process and make the code and model outputs available to ensure transparency and reproducibility in generative model evaluation.

\subsection{Comparing Generative and Loglikelihood-based Evaluation}

A notable advantage of generative evaluation is that it might serve as a better proxy for assessing a language model's performance in real-world applications. In most practical use cases, such as the increasingly popular conversational chatbot format, language models are expected to generate coherent and contextually relevant text based on a given prompt or context. By focusing on the quality and appropriateness of the generated text, generative evaluation provides a more direct assessment of the model's performance in these real-world use cases. This is in contrast to loglikelihood-based tasks, which, while informative, may not fully capture the model's ability to generate text that is both fluent and contextually appropriate.

On the other hand, loglikelihood-based evaluations have their own advantages, particularly when it comes to evaluating smaller or weaker models, or ``base'' models not trained to follow instructions \citep{sanh2022multitask}. These evaluations can provide a useful ranking or measurement of a model's performance, even if the model is not capable of generating high-quality text on its own. By assessing how likely the model is to assign a high probability to the correct answer, loglikelihood-based evaluations can offer insights into the model's understanding of the task. Moreover, techniques like Brier Score can be used to obtain smoother measurements of a model's performance \citep{mirageschaeffer2023emergent}, providing a more nuanced assessment of its capabilities. This can be particularly valuable when comparing and ranking models of different sizes and capacities.




\section{Case Studies}\label{app:case-studies}

\subsection{Comparisons Across Evaluation Settings}\label{app:case-studies-prompt}

Here we provide extra materials and information for the experiment performed in Table \ref{tab:prompt-sensitivity}.

We test on two popular language modeling benchmarks: the ARC question answering benchmark \citep{clark2018think} and MMLU \citep{hendrycks2021measuring}. For several models, we compare how performance on ARC (Challenge subset) and on MMLU varies when we select two distinct settings for each benchmark, varying the prompt (and choice of strings used for loglikelihood-based scoring). Both settings correspond to evaluation setups actually used in the literature at times for these tasks.

ARC was first adapted to the in-context learning setting by \citet{brown2020language} who implement the dataset as a ``cloze'' task: the model is prompted with \texttt{Question: \{question\}$\backslash$nAnswer:} and the likelihood of each potential completion string is compared. By contrast, MMLU \citep{hendrycks2020measuring} provides the model with the question text, each of the 4 possible answers preceded by an answer letter A, B, C, or D, and scores the model based on the generation of the \textit{letter corresponding to the correct answer}. Additionally, \citet{hendrycks2020measuring} aggregate scores via the micro average over all samples instead of the macro average over per-subject scores. However, not all papers evaluate on these tasks in the same way as the original formats. 

However, if models do not adopt these approaches, or disclose their exact settings, it is impossible to reliably compare stated model performance. In Table \ref{tab:prompt-sensitivity}, we compare evaluation on the Challenge 
subset of ARC using the prompt from \citet{brown2020language} (``Cloze'') and using an MMLU-style answer letter with explicit multiple choice options (``MMLU-style''). We additionally compare MMLU scores between the original MMLU prompting style (``MMLU-style'') and an approach we term ``Hybrid'', consisting of an MMLU-style prompt but using the \textit{answer strings} instead of answer letters as the set of continuations over which we can score (this is used by \citet{gpt-neox-20b, penedo2023refinedweb} upon their releases, and in versions of ~\evalharness prior to v0.4.0). As described further below, we can perform this experiment by modifying two lines in \evalharness's ARC and MMLU config files.


We additionally provide the configuration files we use for this experiment, to illustrate an example of the ease of evaluation experimentation, and the configurability and reproducibility afforded. These files' contents can be found below.

All experiments are performed using version 0.4.2 of the LM Evaluation Harness using the \texttt{lm\_eval} PyPI package, which can also be found and downloaded at the following link: \href{https://github.com/EleutherAI/lm-evaluation-harness/tree/v0.4.2/lm\_eval}{https://github.com/EleutherAI/lm-evaluation-harness/tree/v0.4.2/lm\_eval}.

Below are configuration files that may be used to replicate our results on ARC-Challenge and MMLU for all settings, as well as for ARC-Easy (scores not shown).

\lstdefinestyle{yaml}{
     basicstyle=\color{blue}\footnotesize,
     rulecolor=\color{black},
     string=[s]{'}{'},
     stringstyle=\color{blue},
     comment=[l]{:},
     commentstyle=\color{black},
     morecomment=[l]{-}
 }

\begin{lstlisting}[breaklines=True]
group:
  - ai2_arc
task: arc_easy
dataset_path: allenai/ai2_arc
dataset_name: ARC-Easy
output_type: multiple_choice
training_split: train
validation_split: validation
test_split: test
doc_to_text: "Question: {{question}}\nAnswer:"
doc_to_target: "{{choices.label.index(answerKey)}}"
doc_to_choice: "{{choices.text}}"
metric_list:
  - metric: acc
    aggregation: mean
    higher_is_better: true
metadata:
  version: 1.0

\end{lstlisting}

A YAML configuration file for the ARC-easy task, implemented in the ``cloze'' style as done by \citep{brown2020language}.

\begin{lstlisting}[breaklines=true]
group:
  - ai2_arc
task: arc_easy_mmlu
dataset_path: allenai/ai2_arc
dataset_name: ARC-Easy
output_type: multiple_choice
training_split: train
validation_split: validation
test_split: test
doc_to_text: "{{question.strip()}}\n{% for choice in choices.text %}{{choices.label[loop.index - 1]}}. {{choice}}\n{% endfor %}Answer:"
doc_to_target: "{{choices.label.index(answerKey)}}"
doc_to_choice: "{{choices.label}}"
metric_list:
  - metric: acc
    aggregation: mean
    higher_is_better: true
metadata:
  version: 1.0

\end{lstlisting}

A YAML configuration file for the ARC-easy task, as implemented following the prompting style of MMLU in \citet{hendrycks2020measuring}.

We can observe that these configuration files define several components:

\begin{itemize}
    \item The source dataset from the \texttt{Datasets}\citep{lhoest-etal-2021-datasets} library (local datasets are also supported), and the splits to use for testing and few-shot examples. Few-shot examples are drawn from a special fewshot split if specified, else drawn from the training set, validation set, or (in the worst case) non-overlapping test set examples with the current test set example being evaluated, in decreasing order of prioritization.
    \item the \texttt{doc\_to\_*} attributes define mappings to input prompt, gold target label, and the list of answer choice strings, respectively in order.
    \item We provide a list of metrics to use--here, \texttt{acc} denotes unnormalized loglikelihood to score answers, and \texttt{acc\_norm} using byte-length normalization of loglikelihoods.
    \item Finally, the \texttt{metadata.version} field stores the task's version attribute to report.
\end{itemize}

For prototyping, and for the quick modification of interrelated task variants during experimentation, configurations can also inherit from one another: the following is a config file for ARC-challenge, in the cloze style:

\begin{lstlisting}[breaklines=true, ]
include: arc_easy.yaml
task: arc_challenge
dataset_name: ARC-Challenge
\end{lstlisting}

a config file for a single MMLU subset in its original style is the following:

\begin{lstlisting}[breaklines=true, ]
dataset_path: cais/mmlu
test_split: test
fewshot_split: dev
fewshot_config:
  sampler: first_n
output_type: multiple_choice
doc_to_text: "{{question.strip()}}\nA. {{choices[0]}}\nB. {{choices[1]}}\nC. {{choices[2]}}\nD. {{choices[3]}}\nAnswer:"
doc_to_choice: ["A", "B", "C", "D"]
doc_to_target: answer
metric_list:
  - metric: acc
    aggregation: mean
    higher_is_better: true
metadata:
  version: 0.0
\end{lstlisting}

And the "Hybrid" variant:

\begin{lstlisting}[breaklines=true, ]
dataset_path: cais/mmlu
test_split: test
fewshot_split: dev
fewshot_config:
  sampler: first_n
output_type: multiple_choice
doc_to_text: "{{question.strip()}}\nA. {{choices[0]}}\nB. {{choices[1]}}\nC. {{choices[2]}}\nD. {{choices[3]}}\nAnswer:"
doc_to_choice: choices
doc_to_target: answer
metric_list:
  - metric: acc
    aggregation: mean
    higher_is_better: true
metadata:
  version: 0.0
\end{lstlisting}

\section{Best Practices Checklist for Language Model Evaluation}\label{app:best-practices-checklist}

While LM evaluation is difficult, there are measures that can be taken to significantly improve current practices. We provide our high-level recommendations regarding such measures below.

\paragraph{Always share your exact prompts.}
If possible, full \textit{evaluation code} including the full prompts used should be provided for reproducible evaluation runs. Failing this, sharing prompts alone can drastically improve reproducibility.
For fair comparison against other models, evaluation should be done with the same set of prompts unless there is a good reason not to. \textbf{Prompts should not be optimized for performance on a given model, and the amount of prompt engineering done should be disclosed.}

\paragraph{Avoid copying results from other implementations.}
Comparing results across papers can be misleading due to a wide range of experimental differences, including prompts, sample size, metric calculation, and more \citep{marie-etal-2021-scientific}.
Results should \textbf{not} be copied or reported from other papers~\citep{Marie2022-comparing} whenever possible, unless one can verify that the same code has been used to run the experiments in those papers. If such copying is unavoidable, it should be clearly marked as such and treated carefully.

\paragraph{Always provide model outputs.}
Providing model outputs alongside evaluation code can allow others to recalculate scores based on these artifacts, which can be useful for performing statistical significance testing and for assessing the impact of different evaluation metrics or scoring approaches.
Evaluation of large models or APIs can be quite costly---sharing such artifacts allows researchers without access to significant compute to participate in evaluation research.
Finally, sharing outputs can allow results on API models to be reproduced to some extent, even if the models are subsequently deprecated.

\paragraph{Perform qualitative analyses.}
Qualitatively review a small batch of results and outputs before testing at scale: it is very easy to have bugs in your generation code, especially when working with multiple sets of benchmarks, prompts, and models of different architectures. Catching issues early can save a lot of time and compute, and increase confidence in results.
Quantitative scores only provide so much information. To understand why a model is scoring so well or so poorly, it is important to do some sort of qualitative error analysis. This can sometimes reveal superficial errors that are easier to correct with post-processing~\citep{bawden-yvon-bloom-mt-2023}, or more fundamental errors.

\paragraph{Perform statistical significance testing.}
Most works on language modeling do not perform statistical significance testing \citep{marie-etal-2021-scientific}. This simple addition can dramatically boost the reliability of claimed results.
Although costly, reporting results run over more than one random seed can dramatically boost the validity and utility of results, for example by averaging across model runs \citep{multiberts} or averaging over multiple selections of few-shot examples.

\section{Hardware Details}

All experiments are run on 8x NVIDIA A40 48Gb GPUs. Total usage time did not exceed 1 day.

\section{Limitations}\label{app:limitations}

Our work restricts its scope to worries that, and does not discuss issues such as measurement validity in great detail, due to length reasons. There are additionally more case studies and concrete instances of our discussed 

\section{Impacts}\label{app:impacts}

We do not see strong potential negative societal impacts of our work. We believe that healthier and more reliable benchmarking practices and awareness around benchmarking have the potential to improve the societal impacts of LLMs, and hope our recommendations can guide informed and careful assessment of LLMs' capabilities, reducing the odds of their deployment in unsafe and unsuitable scenarios.

\end{document}